\documentclass[reprint, amsmath, amssymb, aps, prl, superscriptaddress]{revtex4-2}

\usepackage{graphicx}
\usepackage{dcolumn}
\usepackage{bm}
\usepackage{hyperref}
\usepackage{multirow}
\usepackage[caption=false]{subfig}
\hypersetup{hidelinks, pdfborder={0 0 0}}
\usepackage{xcolor}

\begin{document}

\preprint{APS/123-QED}

\title{\textbf{Advanced Torrential Loss Function for Precipitation Forecasting}} 

\author{Jaeho Choi}
\email{bearsteak@cau.ac.kr}
\affiliation{AI/ML Innovation Research Center, Chung-Ang University, Seoul 06974, Republic of Korea}

\author{Hyeri Kim}
\email{hyeri.kim@ou.edu}
\affiliation{Advanced Radar Research Center, University of Oklahoma, Norman, OK 73019, USA}
\affiliation{School of Meteorology, University of Oklahoma, Norman, OK 73072, USA}

\author{Kwang-Ho Kim}
\email{khkim777@korea.kr}
\affiliation{Weather Radar Center, Korea Meteorological Administration, Seoul 07062, Republic of Korea}

\author{Jaesung Lee}
\thanks{Corresponding author:~\href{mailto:curseor@cau.ac.kr}{curseor@cau.ac.kr}\\}
\affiliation{AI/ML Innovation Research Center, Chung-Ang University, Seoul 06974, Republic of Korea}
\affiliation{Department of Artificial Intelligence, Chung-Ang University, Seoul 06974, Republic of Korea}


\date{\today}

\begin{abstract} 
Accurate precipitation forecasting is becoming increasingly important in the context of climate change.
In response, machine learning-based approaches have recently gained attention as an emerging alternative to traditional methods such as numerical weather prediction and climate models.
Nonetheless, many recent approaches still rely on off-the-shelf loss functions, and even the more advanced ones merely involve optimization processes based on the critical success index (CSI).
The problem, however, is that CSI may become ineffective during extended dry periods when precipitation remains below the threshold, rendering it less than ideal as a criterion for optimization.
To address this limitation, we introduce a simple penalty expression and reinterpret it as a quadratic unconstrained binary optimization (QUBO) formulation.
Ultimately, the resulting QUBO formulation is relaxed into a differentiable advanced torrential (AT) loss function through an approximation process.
The proposed AT loss demonstrates its superiority through the Lipschitz constant, forecast performance evaluations, consistency experiments, and ablation studies with the operational model.
\end{abstract}

\maketitle

\emph{Introduction.---}Climate change has altered weather patterns, giving rise to meteorological phenomena not previously observed in certain regions~\cite{ghil2020physics, stott2016climate}.
Extreme weather events are being recorded with increasing frequency worldwide, and the adverse impacts have intensified, especially from localized torrential rainfall and extended dry periods~\cite{cutillas2023local, kim2025observational, francis2025cause}.
These events, by inducing frequent outliers, underscore the need to maintain reliable precipitation forecasting performance under such conditions.
In this context, the critical success index (CSI) has come to be regarded as the metric most often considered first in evaluating precipitation forecasting~\cite{schaefer1990critical, doswell1990summary}.
In practice, recently emerging machine learning (ML)-based approaches use CSI as the criterion to optimize forecast quality.
They achieve this either by employing complex architectures or by simply incorporating CSI as a constraint term in objective functions with multiple loss functions~\cite{andrychowicz2023deep, zhang2023skilful, choi2023pct}.
Nevertheless, the primary loss functions, which remain the centerpiece of optimization, still rely on off-the-shelf designs, and even the suitability of CSI as a criterion for optimization remains debated.

In this Letter, we propose an advanced torrential (AT) loss function for broad use in ML-based precipitation forecasting, following the identification of the mathematical limitations of the CSI as an optimization criterion and addressing them through a binary penalty.
The binary penalty is computed for each grid cell, equivalent to a pixel in the image domain, and the aggregated penalty forms the foundation of the AT loss.
Furthermore, the overall penalty is expressed in the quadratic unconstrained binary optimization (QUBO) formulation and approximated as a differentiable surrogate using the binary Gumbel-Softmax, forming the AT loss that converges toward minimizing the penalty~\cite{kochenberger2014unconstrained, gumbel1941return}.
The AT loss is validated in three key aspects.
The first aspect is training stability, ensured by a low Lipschitz constant in the early stage of training.
The second aspect is forecast performance, evaluated by how accurately precipitation greater than or equal to the operational threshold is simulated across lead times, with outperformance confirmed via various forecasting metrics.
In particular, forecast performance is validated via a convolutional long short-term memory (ConvLSTM) encoder-decoder model, which serves as a widely used benchmark in precipitation forecasting~\cite{shi2015convolutional}.
The third aspect is performance consistency under outliers, which can be induced by extreme weather events, with its superiority confirmed through comparisons with commonly used pixel-wise loss functions.
From a practical standpoint, the effectiveness of the AT loss is further demonstrated through ablation studies using the paired complementary temporal cycle-consistent adversarial networks (PCT-CycleGAN)~\cite{choi2023pct}.

For the convenience of scholars from diverse fields, a glossary is also provided in the End Matter.

\emph{Characteristics of CSI.---}Given real precipitation data $X^{t}=(x_i^{t})_{i=1}^{n}$ from $n$ grid cells at time step $t$, the ML-based forecasting model is trained to generate forecast data $Y^{t+1}=(y_i^{t+1})_{i=1}^{n}$ that closely approximates the ground-truth $X^{t+1}$.
In particular, recent advanced models aim to make the CSI of $Y^{t+1}$ as close to $1$ as possible through iterative training~\cite{choi2023pct, andrychowicz2023deep}.
This metric, denoted as $\mathrm{CSI}(Y^{t+1})$ is computed as
\begin{align}
    \mathrm{CSI}(Y^{t+1}) &= \frac{\sum_{i=1}^{n} \left( f(x_i^{t+1}) \land f(y_i^{t+1}) \right)}{\sum_{i=1}^{n} \left( f(x_i^{t+1}) \lor f(y_i^{t+1}) \right)}, \nonumber \\
    &\text{where } f(k) =
    \begin{cases}
        1, & \text{if } k \ge \theta, \\
        0, & \text{otherwise}.
    \end{cases}
\label{eq:csi}
\end{align}
Here, $f(k)$ is a step function and $\theta\in\mathbb{R}_{\geq 0}$ denotes the threshold.
According to Eq.~(\ref{eq:csi}), when all values fall below the threshold, the denominator of $\mathrm{CSI}(Y^{t+1})$ becomes $0$, making it unreliable during extended dry periods with negligible or no precipitation.
Furthermore, the numerator of $\mathrm{CSI}(Y^{t+1})$ excludes any grid cell $i$ where both $f(x_i^{t+1})$ and $f(y_i^{t+1})$ are $0$, even though such cases represent correct forecasts.
In addition to these characteristics, the presence of $f(k)$ makes $\mathrm{CSI}(Y^{t+1})$ non-differentiable and therefore cannot be directly optimized with gradient descent~\cite{costa2022deep, klos2025smooth}.
Thus, the CSI is suboptimal as a training criterion.
Nonetheless, since the CSI is effective for evaluating precipitation forecasts, an advanced optimization criterion that builds on its strengths while addressing its limitations is required.

\begin{table}[t]
    \caption{Binary penalty for four distinct cases.}
    \label{tab:binary_penalty}
    \begin{ruledtabular}
        \begin{tabular}{c|cccc}
            Case & $f(x_i^{t+1})$ & $f(y_i^{t+1})$ & Forecast & Penalty \\
            \colrule
            1 & 0 & 0 & $\mathrm{True}$ & 0 \\
            2 & 0 & 1 & $\mathrm{False}$ & 1\\
            3 & 1 & 0 & $\mathrm{False}$ & 1\\
            4 & 1 & 1 & $\mathrm{True}$ & 0\\
        \end{tabular}
    \end{ruledtabular}    
\end{table}
\emph{Binary Penalty.---}Based on Eq.~(\ref{eq:csi}), the forecast at grid cell $i$ for time step $t+1$ can be divided into four distinct cases, as shown in Table~\ref{tab:binary_penalty}.
In case~$1$, both the real value $x_i^{t+1}$ and the generated forecast value $y_i^{t+1}$ are below the threshold $\theta$, meaning that $f(x_i^{t+1})=f(y_i^{t+1})=0$.
In this case, the forecast outcome at grid cell $i$ cannot affect $\mathrm{CSI}(Y^{t+1})$.
Since this point gives rise to the limitations of CSI, a binary penalty is introduced as a straightforward solution.
The penalty is applied only to incorrect forecasts that are considered false.
Thus, cases~$1$ and $4$, where $f(x_i^{t+1})=f(y_i^{t+1})$, are true forecasts with no penalty, while cases~$2$ and $3$, where $f(x_i^{t+1})\neq f(y_i^{t+1})$, are false forecasts with a penalty of $1$.
Ultimately, the binary penalty $P_i^{t+1}$ at grid cell $i$ is defined as
\begin{align}
    P_i^{t+1} = f(x_i^{t+1}) + f(y_i^{t+1}) - 2f(x_i^{t+1})f(y_i^{t+1}).
\label{eq:p_i}
\end{align}
The overall penalty $P^{t+1}$, which is the sum of $P_i^{t+1}$, simplifies to
\begin{align}
     P^{t+1} = \sum_{i=1}^{n} \left( f(x_i^{t+1}) - f(y_i^{t+1}) \right)^{2}. 
\label{eq:penalty}
\end{align}
Since improving forecast performance requires minimizing Eq.~(\ref{eq:penalty}), this expression naturally serves as the objective function of the QUBO formulation and a potential criterion for optimizing ML-based approaches~\cite{albash2018adiabatic, farhi2014quantum, wang2018quantum}.

\emph{Approximation.---}The most suitable approach to using Eq.~(\ref{eq:penalty}) in optimizing ML-based forecasting models is to let it serve as the foundation of the loss function.
This is achieved by approximating only $f(y_i^{t+1})$, since $f(x_i^{t+1})$ is fixed by the constant $x_i^{t+1}$.
To this end, $f(y_i^{t+1})$ can be approximated via binary Gumbel-Softmax, a differentiable relaxation of Bernoulli sampling~\cite{maddison2014sampling, jang2016categorical}.
This method is stable and can sharply discriminate the threshold, allowing accurate identification of the presence or absence of precipitation when integrated with the QUBO formulation.
In this context, the logits for $1$ and $0$ are represented as $l_1 = y_i^{t+1} - \theta$ and $l_0 = -l_1$, respectively, depending on the threshold $\theta$.
The resulting approximation $\tilde f(y_i^{t+1})$ is defined as
\begin{align}
    \tilde f(y_i^{t+1}) &= \frac{1}{1 + \exp \left(-\frac{1}{\tau} \left( l_1 - l_0 + g_1 - g_0 \right) \right)} \nonumber \\
    &= \sigma \left( \frac{1}{\tau} \left( 2y_i^{t+1} - 2\theta + z \right) \right),
\label{eq:f_app}
\end{align}
where $g_0,~g_1 \sim \mathrm{Gumbel}(0,~1)$ and $z \sim \mathrm{Logistic}(0,~1)$.
Note that $\tau > 0$ is the temperature parameter and $\sigma(\cdot)$ is the sigmoid function.

\emph{AT Loss.---}According to Eqs.~(\ref{eq:penalty})--(\ref{eq:f_app}), the mean of the approximation of $P^{t+1}$ can be used as the loss function.
Therefore, the proposed AT loss between the ground-truth $X^{t+1}$ and the forecasts $Y^{t+1}$ is defined as
\begin{align}
    \mathcal{L}&(X^{t+1},~Y^{t+1};~\tau,~\theta,~z) \nonumber \\ 
    &= \frac{1}{n} \sum_{i=1}^{n} \mathcal{L}_i \nonumber \\
    &= \frac{1}{n} \sum_{i=1}^{n} \left( f(x_i^{t+1}) - \sigma \left( \frac{1}{\tau} \left( 2y_i^{t+1} - 2\theta + z \right) \right) \right)^{2},
\label{eq:at_loss}
\end{align}
where $\mathcal{L}_i$ is the individual loss for grid cell $i$.
Here, $z$ is obtained through inverse transform sampling as $\ln u - \ln (1 - u)$, where $u \sim \mathrm{Uniform}(0,~1)$.
In the actual implementation, $\tau \in (0,~1]$ is gradually decayed across training epochs, and the perturbation is constrained by $\lvert z \rvert \ll 1$.
This strategy helps ensure stable optimization while leveraging a proven annealing-based hyperparameter control technique~\cite{jang2016categorical, jo2024annealing}.
Another point to note is that, as the AT loss already involves $\sigma(\cdot)$, the activation function at the output layer should be omitted, or the pre-activation values should be used, to avoid information loss from redundant non-linear transformations.

\emph{Training Stability.---}According to Eq.~(\ref{eq:at_loss}), differentiating $\mathcal{L}_i$ with respect to $y_i^{t+1}$ yields
\begin{align}
    \frac{ \partial \mathcal{L}_i}{ \partial y_i^{t+1} \hspace{-7pt}} ~&= -\frac{4}{\tau} \left( f(x_i^{t+1}) - \zeta \right) \zeta \left( 1 - \zeta \right), \nonumber \\
    &~\text{where } \zeta = \sigma \left( \frac{1}{\tau} \left( 2y_i^{t+1} - 2\theta + z \right) \right).
\label{eq:dl_dy}
\end{align}
Since $f(x_i^{t+1}) \in \{0,~1\}$, Eq.~(\ref{eq:dl_dy}) simplifies as
\begin{align}
    \frac{ \partial \mathcal{L}_i}{ \partial y_i^{t+1} \hspace{-7pt}} ~=
    \begin{cases}
        -\frac{4}{\tau} \left( \zeta^{3} - \zeta^{2} \right), & \text{if } f(x_i^{t+1}) = 0, \\
        -\frac{4}{\tau} \left( \zeta^{3} - 2\zeta^{2} + \zeta \right), & \text{otherwise}.
    \end{cases}
\label{eq:cases_dl_dy}
\end{align}
Differentiating Eq.~(\ref{eq:cases_dl_dy}) with respect to $\zeta$ simplifies to
\begin{align}
    \frac{\partial}{\partial \zeta} \left( \frac{\partial \mathcal{L}_i}{\partial y_i^{t+1} \hspace{-7pt}} ~\right) =
    \begin{cases}
        -\frac{4}{\tau} \zeta \left( 3\zeta - 2 \right), & \text{if } f(x_i^{t+1}) = 0, \\
        -\frac{4}{\tau} \left( \zeta - 1 \right) \left( 3\zeta - 1 \right), & \text{otherwise}.
    \end{cases}
\label{eq:dzeta_dl_dy}
\end{align}
Thus, the extremum of Eq.~(\ref{eq:cases_dl_dy}) is attained when $\zeta = \frac{2}{3}$ for $f(x_i^{t+1}) = 0$, and when $\zeta = \frac{1}{3}$ for $f(x_i^{t+1}) = 1$.
Substituting the value of $\zeta$ for each case, the Lipschitz constant of $\mathcal{L}_i$ is obtained as $\frac{16}{27\tau} \approx 0.5926\tau^{-1}$.
In practice, the fact that the Lipschitz constant remains below $1$ in the interval $\tau \in [0.6,~1]$, where most of the optimization takes place, guarantees the stability of the training process.

\emph{Data and Settings.---}The AT loss was evaluated based on training and test datasets derived from composite hybrid surface rainfall (HSR) data of the Korea Meteorological Administration (KMA) used in official weather forecasting~\cite{choi2023pct, tsai2025association}.
Each instance of HSR data represents the precipitation intensity over a $1024$~km $\times$ $1024$~km domain covering the entire Republic of Korea.
The training dataset was constructed from HSR data spanning $2022$ to $2023$, whereas the test dataset was constituted from data collected between January and August $2024$, together with data obtained in July $2025$.
Both datasets were specified at spatial and temporal resolutions of $4$ km and $10$ minutes, respectively.
In particular, to preserve temporal causality, the datasets were organized using a sliding-window approach, where each window consisted of six consecutive time steps~\cite{gao2014transmission, ozken2015transformation}.
The datasets were normalized to the range $[-1,~1]$.
Details of hyperparameter settings are provided in the End Matter.

\emph{Baselines.---}The AT loss was compared with various pixel-wise loss functions.
The mean absolute error (MAE) and mean squared error (MSE) losses, widely used in the field, were adopted as baselines~\cite{eskicioglu1995image}.
Additionally, the Huber and Charbonnier losses, which combine the advantages of the previous two losses while providing robustness, were also included as baselines~\cite{huber1992robust, charbonnier1994two, lai2017deep}.

\begin{table}[t]
    \caption{Forecast performance comparisons. The row and column headers indicate loss functions and metrics, respectively. All abbreviations are defined in the text.}
    \label{tab:convlstm}
    \begin{ruledtabular}
        \begin{tabular}{c|cccc}
            Loss & \multicolumn{1}{c}{CSI} & \multicolumn{1}{c}{HSS} & \multicolumn{1}{c}{POD} & \multicolumn{1}{c}{FAR} \\
            \colrule
            & \multicolumn{4}{c}{$2$-step forecast ($20$-minute)} \\
            AT & \textbf{0.6015} & \textbf{0.7478} & 0.7173 & \textbf{0.2117} \\
            MAE & 0.5618 & 0.7165 & 0.6658 & 0.2174 \\
            MSE & 0.5055 & 0.6673 & \textbf{0.7538} & 0.3945 \\
            Huber & 0.4375 & 0.6047 & 0.5431 & 0.3077 \\
            Charbonnier & 0.5702 & 0.7231 & 0.7146 & 0.2616 \\
            & \multicolumn{4}{c}{$4$-step forecast ($40$-minute)} \\
            AT & \textbf{0.4980} & \textbf{0.6606} & 0.6093 & \textbf{0.2684} \\
            MAE & 0.4590 & 0.6253 & 0.5626 & 0.2865 \\
            MSE & 0.4134 & 0.5792 & \textbf{0.7082} & 0.5018 \\
            Huber & 0.3746 & 0.5402 & 0.4974 & 0.3973 \\
            Charbonnier & 0.4612 & 0.6273 & 0.5895 & 0.3206 \\
            & \multicolumn{4}{c}{$6$-step forecast ($60$-minute)} \\
            AT & \textbf{0.4172} & \textbf{0.5838} & 0.5176 & \textbf{0.3174} \\
            MAE & 0.3830 & 0.5495 & 0.4759 & 0.3377 \\
            MSE & 0.3507 & 0.5124 & \textbf{0.6664} & 0.5747 \\
            Huber & 0.3386 & 0.5000 & 0.5140 & 0.5020 \\
            Charbonnier & 0.3798 & 0.5459 & 0.4908 & 0.3732 \\
        \end{tabular}            
    \end{ruledtabular}    
\end{table}
\emph{Forecast Evaluations with Benchmark.---}Comparative experiments were conducted on the AT loss using the ConvLSTM encoder-decoder model, a benchmark in ML-based operational precipitation forecasting systems~\cite{shi2015convolutional}.
The evaluation adopted four recommended metrics for weather forecasting verification: CSI, Heidke skill score (HSS), probability of detection (POD), and false alarm ratio (FAR)~\cite{nurmi2003recommendations, das2024hybrid}.
Better forecasting performance is indicated when FAR approaches $0$, while the other metrics are better when closer to $1$.

According to Table~\ref{tab:convlstm}, the AT loss achieved the best CSI, HSS, and FAR across all lead times of $20$, $40$, and $60$ minutes, outperforming the other baselines.
Although the MSE loss showed the highest POD, this came with the highest FAR.
Since effective forecasts require both high POD and low FAR, the AT loss provided the best overall balance, demonstrating superior performance.

\begin{figure}[t]
    \centering
    \includegraphics[width=0.842\linewidth]{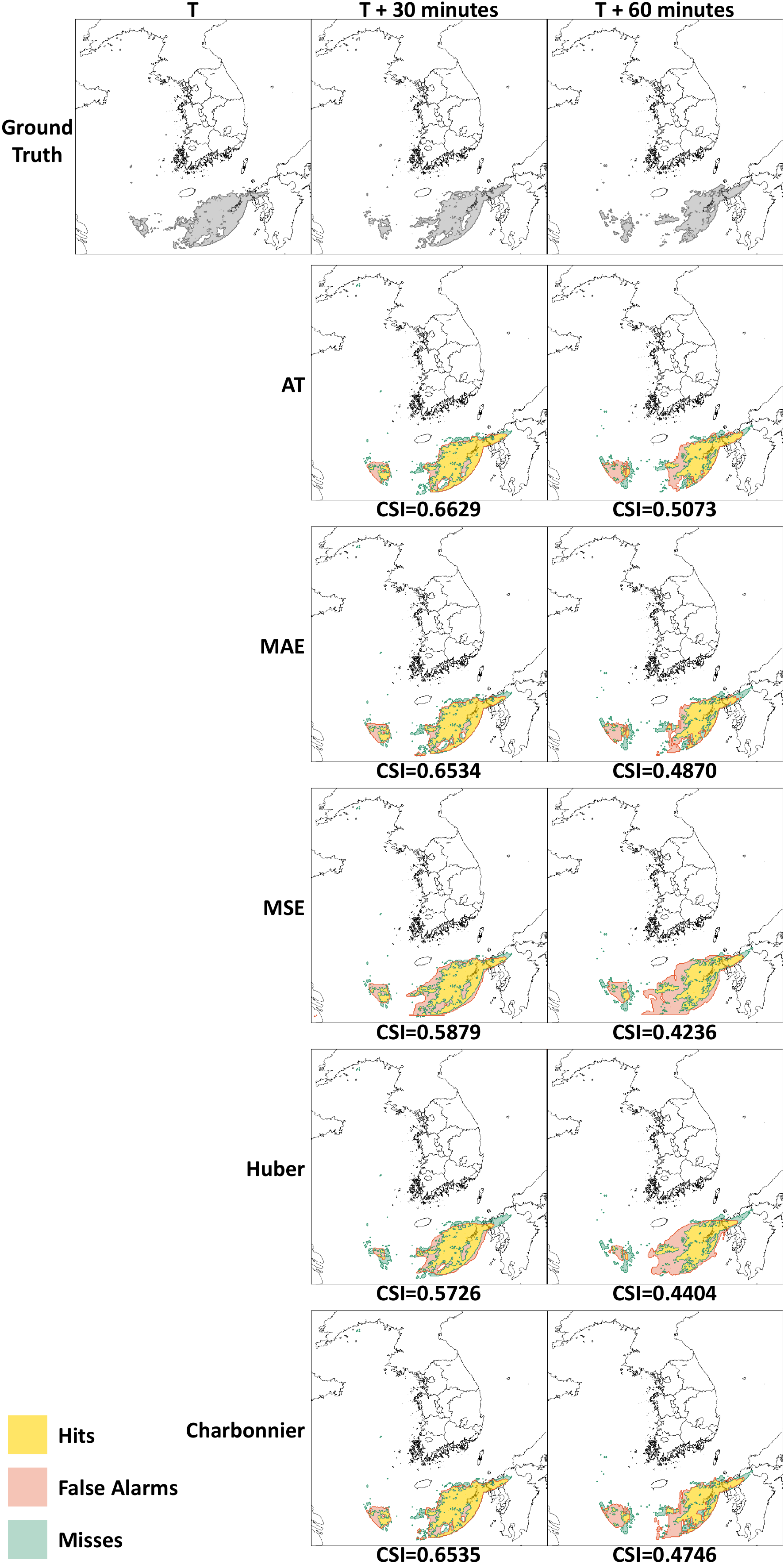}
    \caption{A case of precipitation forecast comparisons from time $\mathrm{T}$ (2024-06-26 01:50:00 UTC).}
    \label{fig:convlstm_case}
\end{figure}
To enable a more intuitive comparison, Fig.~\ref{fig:convlstm_case} illustrates a case of $1$-hour forecasts at 01:50:00 UTC on June $26$, $2024$.
The highlighted areas represent regions with precipitation intensity greater than or equal to $2$~mm/h.
In this case, the AT loss produced the best forecasts beginning from the $30$-minute lead time.
In particular, it most accurately simulated the eastward-moving precipitation echoes.
As a supplement, Figs.~\ref{fig:convlstm_case_extended} and \ref{fig:convlstm_appendix} in the End Matter present extended results of Fig.~\ref{fig:convlstm_case} and an additional case, respectively.

\emph{Consistency Experiments under Outliers.---}The performance consistency of the AT loss under outliers was verified through comparative experiments with baselines.
To use only the intended outliers in the experiments, the interquartile range of the original data was first calculated, and Tukey's method was applied to identify the existing outliers within the range~\cite{tukey1977exploratory}.
The identified outliers were replaced with the mean precipitation intensity of neighboring grid cells.
Then, this refined data was used to train a convolutional neural network (CNN) for $1$-step forecasting in two tracks, clean and dirty, for each loss function, with architectural details provided in the End Matter.
The clean track used the refined data as is for training.
In the dirty track, intentional noise was injected as the refined data was fed into the CNN.
Specifically, $10$ to $30$ percent of the refined data was randomly corrupted with salt-and-pepper noise and random-valued impulse noise~\cite{zhao2022noise, dong2007detection}.
After training, forecasts from both tracks were compared in terms of MAE and peak signal-to-noise ratio (PSNR)~\cite{hore2010image, liu2025active}.
The lower the MAE and the higher the PSNR between the forecasts from the clean and dirty tracks, the more consistent the performance of the loss function is, regardless of contamination.

\begin{table}[t]
    \caption{Performance differences under outliers. The row and column headers indicate loss functions and metrics, respectively. All abbreviations are defined in the text.}
    \label{tab:fcnn}
    \begin{ruledtabular}
        \begin{tabular}{c|cc}
            Loss & \multicolumn{1}{c}{MAE} & \multicolumn{1}{c}{PSNR} \\
            \colrule
            & \multicolumn{2}{c}{Salt-and-pepper noise} \\
            AT & \textbf{2.4251} & \textbf{33.8598} \\
            MAE & 4.0917 & 33.4847 \\
            MSE & 8.9409 & 27.4833 \\
            Huber & 4.2883 & 33.4819 \\
            Charbonnier & 4.5570 & 33.2172 \\
            & \multicolumn{2}{c}{Random-valued impulse noise} \\
            AT & \textbf{2.9976} & \textbf{34.0679} \\
            MAE & 5.3458 & 32.2673 \\
            MSE & 11.1006~\hspace{0.5mm} & 25.4866 \\
            Huber & 4.9959 & 33.0321 \\
            Charbonnier & 5.4007 & 32.0984 \\
        \end{tabular}
    \end{ruledtabular}    
\end{table}
As presented in Table~\ref{tab:fcnn}, the AT loss achieved the lowest MAE and the highest PSNR across both noise types, clearly outperforming the baselines.
It is particularly encouraging that the performance gap was more pronounced under random-valued impulse noise, which better reflects realistic scenarios.
These results suggest that the AT loss enables stable training, allowing forecasting models to maintain consistent performance in the presence of outliers induced by extreme weather events.

\emph{Ablation Studies.---}To evaluate the contribution of AT loss to forecast performance in operational practice, ablation studies were conducted using the $2025$ version of the PCT-CycleGAN~\cite{choi2023pct}.
This version, adapted from the initial model, replaced shot-to-shot processing with a sequence-to-sequence framework and upgraded the torrential loss into AT loss.
All other experimental conditions remained identical, with the only difference being the presence or absence of AT loss.
For reproducibility, the ablation studies utilized the KMA color map and applied an operational CSI threshold of $0.5$~mm/h.

\begin{figure}
    \centering
    \subfloat[Case forecast initialized at time $\mathrm{T}$ (2024-01-10 00:00:00 UTC).]{
        \includegraphics[width=0.9769\linewidth]{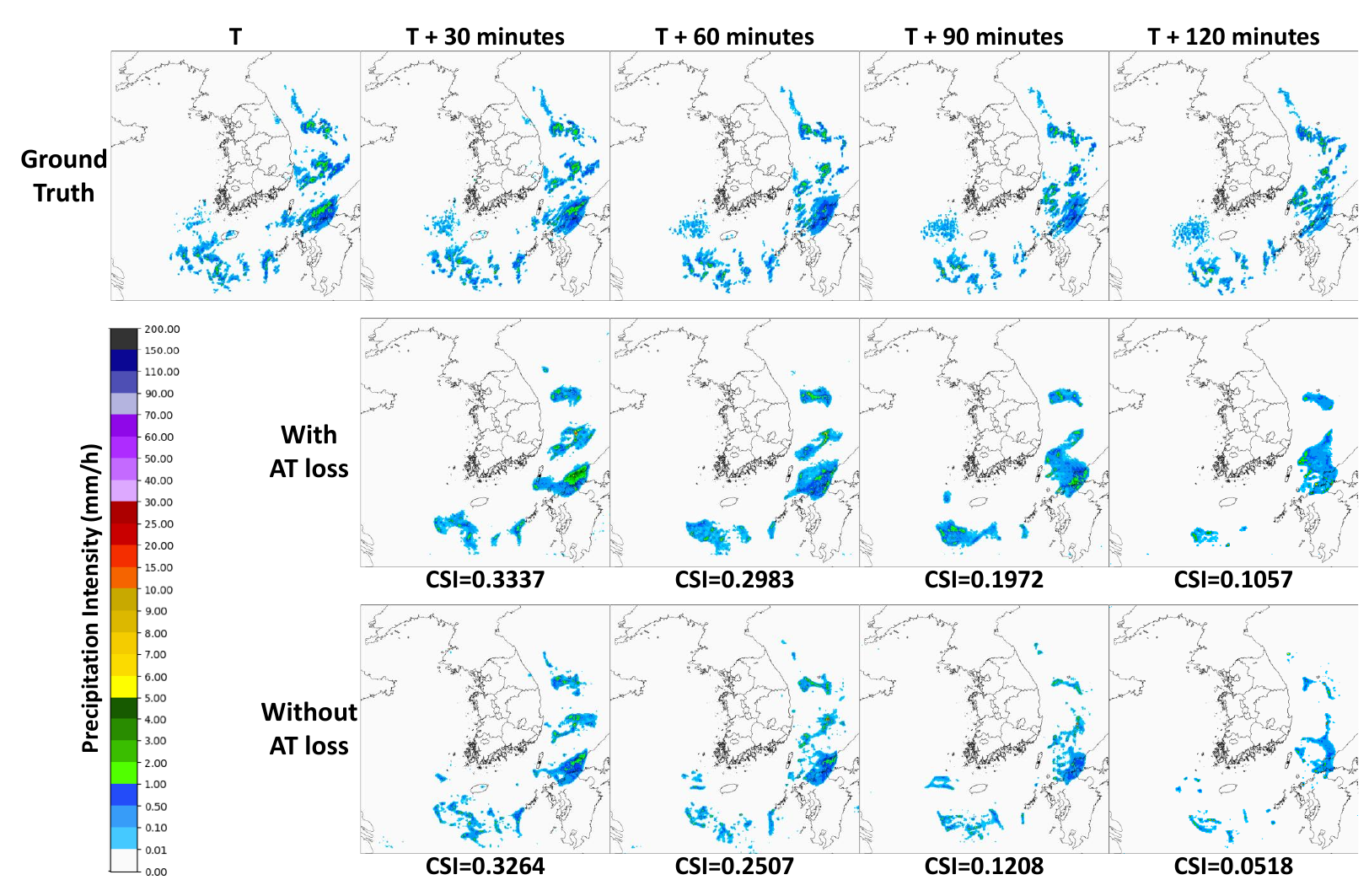}
        \label{fig:pct_light}}\\
    \subfloat[Case forecast initialized at time $\mathrm{T}$ (2025-07-16 15:00:00 UTC).]{
        \includegraphics[width=0.9769\linewidth]{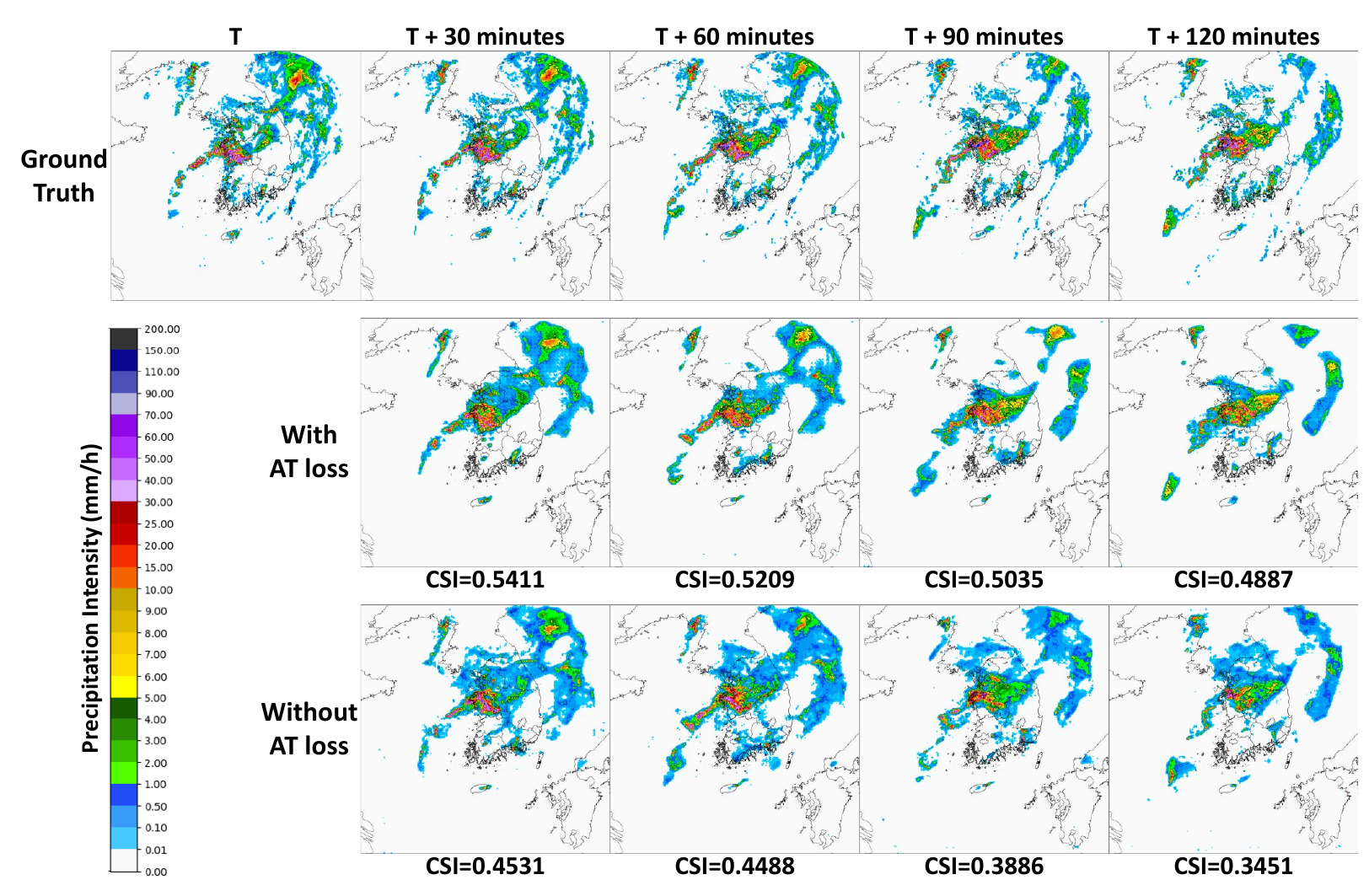}
        \label{fig:pct_heavy}}
    \caption{Forecasts with and without advanced torrential (AT) loss in ablation studies on light and heavy precipitation cases.}
\end{figure}
Fig.~\ref{fig:pct_light} presents $2$-hour forecasts of a light precipitation case.
The initial time $\mathrm{T}$ is 00:00:00 UTC on January $10$, $2024$.
In this case, the threshold for AT loss was set to $0.5$~mm/h, with a focus on light precipitation.
For all lead times, the model with AT loss outperformed the one without it in terms of the CSI metric, and the gap between the two grew over time.
In the final lead time, the model with AT loss achieved more than twice the CSI of the model without it.
This suggests a positive effect on improving forecasting performance for light precipitation, which is more challenging than typical cases.

Fig.~\ref{fig:pct_heavy} shows the $2$-hour forecasts for a heavy precipitation case starting at 15:00:00 UTC on July $16$, $2025$.
This case represents an exceptionally intense regional precipitation event.
The model with AT loss outperformed the one without it across all lead times, and at the last lead time, it achieved approximately $42$ percent higher CSI.
From a qualitative perspective, meteorologists at the KMA judged that the model incorporating AT loss provided a more accurate simulation of heavy precipitation areas than the model without it.

As an extension of these ablation studies, the End Matter discusses in detail the considerations for applying the AT loss to worldwide operational models used in professional institutions.

\emph{Conclusion.---}In this study, we introduced a grid cell-level binary penalty to overcome the limitations of CSI.
On this basis, we proposed the AT loss, formulated with QUBO and the binary Gumbel-Softmax, to advance ML-based forecasting.
The AT loss was designed to ensure stable convergence in early training and sharp threshold discrimination in later stages.
Experiments showed that AT loss delivered the best forecast performance and was consistent under outliers, with its effectiveness further supported by ablation studies in operational settings.
Nevertheless, broader validation across operational models in different countries remains a common challenge for the atmospheric physics and meteorology communities.

\begin{acknowledgments}
\emph{Acknowledgments.---}The authors thank Yura Kim and Seongmook Kim for insightful discussions.
This work was supported by the National Research Foundation of Korea (NRF) grant funded by the Korea government (MSIT) (No. RS-2024-00459387), and the Development of Integrated Application Technology for Korea Weather Radar project funded by the Weather Radar Center, Korea Meteorological Administration [KMA2021-03122, Development of Radar Based Severe Weather Nowcasting Technology].

\emph{Code Availability.---}The official implementation of the AT loss is publicly available at~\url{https://github.com/kaetsraeb/at-loss}.

\emph{Data Availability.---}The data that support the findings of this work are openly available~\cite{data2025kma}.
\end{acknowledgments}

\bibliography{apssamp}

\appendix
\section*{End Matter}

\emph{Glossary.---}To accommodate a diverse readership, concise explanations of key terms are provided below.
All terms are listed alphabetically.
For clarity, abbreviations are not used in the glossary descriptions.

\noindent
\raisebox{0.3ex}{\scalebox{0.6}{$\bullet$}}\ Bernoulli sampling: a random binary sampling process where each element is independently assigned the value $1$ with probability $p$ and $0$ with $1-p$.

\noindent
\raisebox{0.3ex}{\scalebox{0.6}{$\bullet$}}\ False alarm ratio: the ratio of false alarms, where the event was forecast but not observed, to the total number of forecasts of the event.

\noindent
\raisebox{0.3ex}{\scalebox{0.6}{$\bullet$}}\ Gumbel-Softmax: a differentiable reparameterization of categorical sampling that combines Gumbel noise with a softmax-based relaxation, enabling gradient-based optimization for models involving discrete variables. 
In the binary case, it provides a continuous relaxation of the Bernoulli distribution.

\noindent
\raisebox{0.3ex}{\scalebox{0.6}{$\bullet$}}\ Heidke skill score: a measure of forecast accuracy that compares the proportion of correct forecasts to that expected by random chance, indicating the improvement over random guessing.
Values below $0$ indicate performance worse than random chance, $0$ indicates no skill, and $1$ represents a perfect forecast.

\noindent
\raisebox{0.3ex}{\scalebox{0.6}{$\bullet$}}\ Lead time: the time interval, in meteorology and hydrology, between the issuance of a forecast and the expected occurrence of the forecast event.

\noindent
\raisebox{0.3ex}{\scalebox{0.6}{$\bullet$}}\ Logit: the logarithm of the odds of a probability, often used to transform probabilities into unbounded real values.
In machine learning, it denotes the value before conversion into a probability by the activation function.

\noindent
\raisebox{0.3ex}{\scalebox{0.6}{$\bullet$}}\ Probability of detection: the ratio of correctly forecast events to the total number of observed events, indicating how often observed events were successfully forecast.

\noindent
\raisebox{0.3ex}{\scalebox{0.6}{$\bullet$}}\ Quadratic unconstrained binary optimization: a mathematical formulation for solving combinatorial optimization problems, where a quadratic objective function is minimized over binary variables without explicit constraints. 
It serves as the standard problem representation for quantum optimization.

\noindent
\raisebox{0.3ex}{\scalebox{0.6}{$\bullet$}}\ Salt-and-pepper noise: a type of impulse noise where random pixels are abruptly set to either the maximum (salt) or minimum (pepper) intensity.

\noindent
\raisebox{0.3ex}{\scalebox{0.6}{$\bullet$}}\ Tukey's method: a statistical technique for identifying outliers based on the interquartile range, where data points lying beyond $1.5$ times the interquartile range from the first or third quartile are considered outliers.

\emph{Hyperparameter Settings.---}In the experiments conducted with different objectives, the following common conditions were applied unless otherwise specified.
Adam was used as the optimizer with momentum parameters $\beta_1=0.9$ and $\beta_2=0.999$~\cite{kingma2014}.
The batch size and learning rate were set to $16$ and $0.0002$, respectively.
Instance normalization and the Swish activation function were incorporated into the hidden layers~\cite{ulyanov2017improved, ramachandran2017searching}.
The default threshold of $2$~mm/h was applied for the AT loss and forecasting metrics.
The value of the decaying $\tau$ in Eq.~(\ref{eq:at_loss}) was limited to a minimum of $0.05$.

\begin{figure}[b!]
    \centering
    \includegraphics[width=\linewidth]{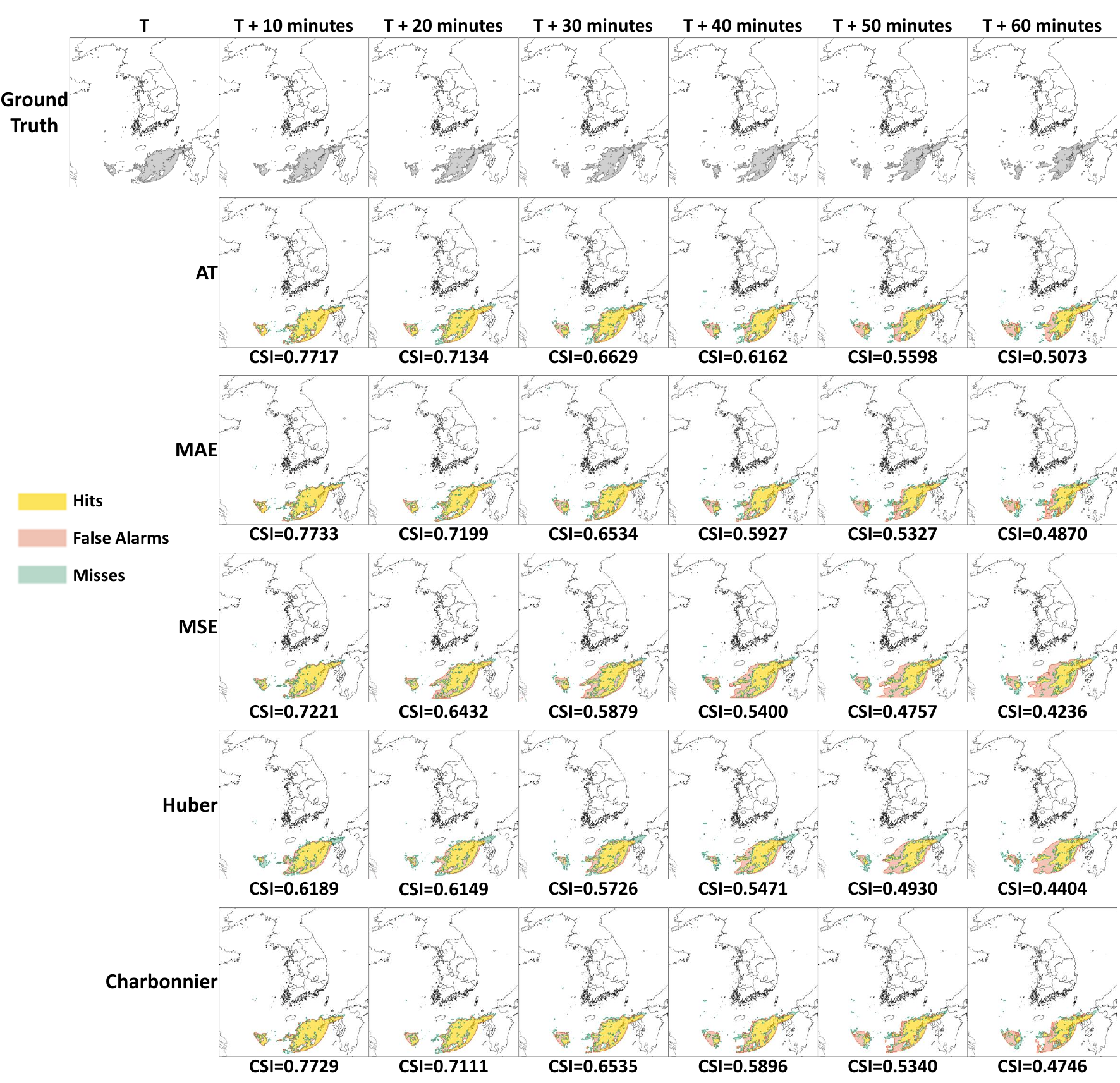}
    \caption{An extended case of precipitation forecast comparisons from time $\mathrm{T}$ (2024-06-26 01:50:00 UTC).}
    \label{fig:convlstm_case_extended}
\end{figure}

\emph{Extended Results of Fig.~\ref{fig:convlstm_case}.---}Fig.~\ref{fig:convlstm_case_extended} presents forecast comparisons for all lead times at $10$-minute intervals.

\begin{figure}[b!]
    \centering
    \includegraphics[width=\linewidth]{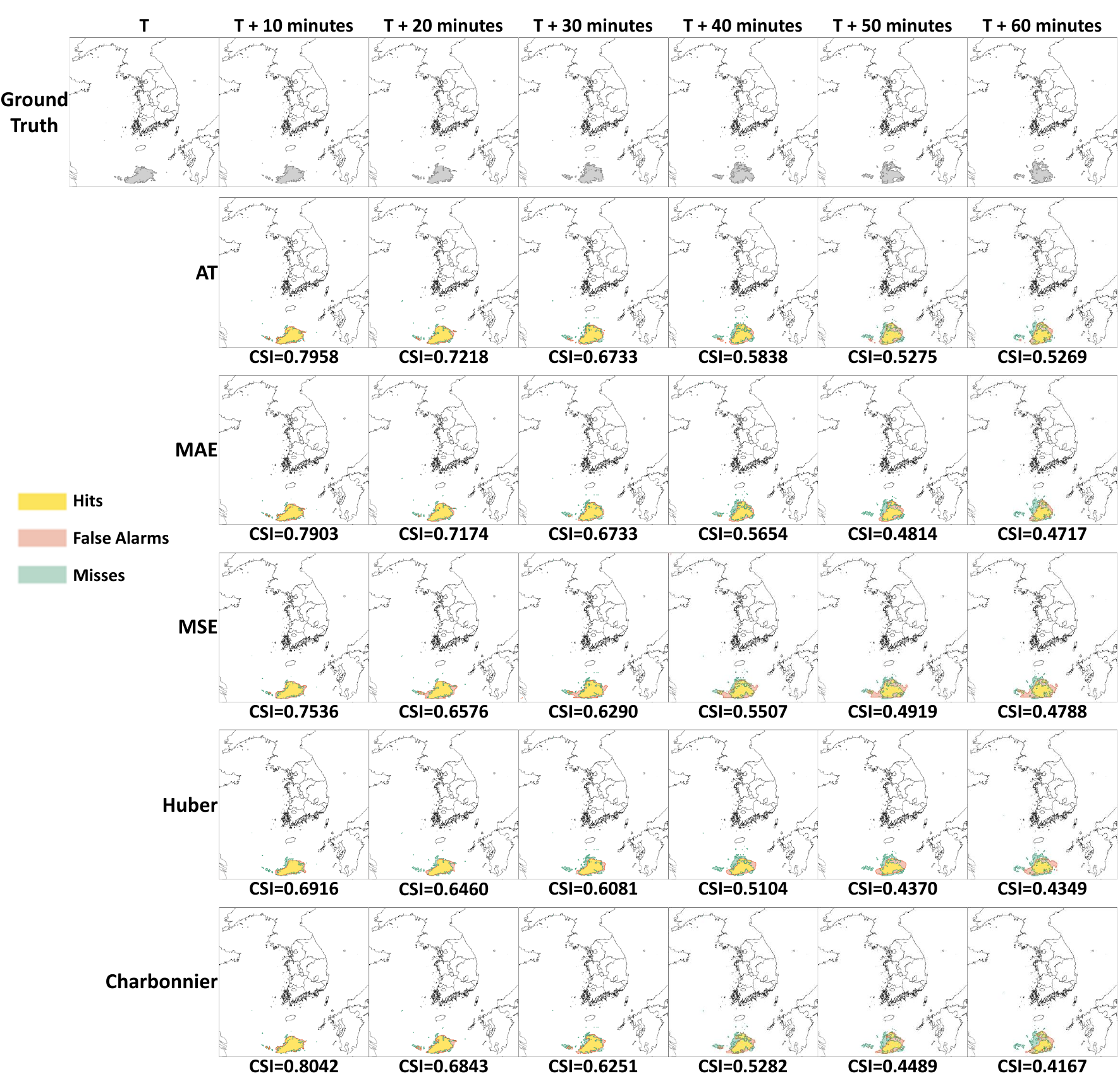}
    \caption{A case of precipitation forecast comparisons from time $\mathrm{T}$ (2024-06-05 17:10:00 UTC).}
    \label{fig:convlstm_appendix}
\end{figure}

\emph{Additional Forecast Comparisons.---}An additional case of $1$-hour forecasts, issued at 17:10:00 UTC on June $5$, $2024$, is shown in Fig.~\ref{fig:convlstm_appendix}.
The AT loss produced the best forecasts from the $20$-minute lead time onward, recording the highest CSI.
In particular, it simulated the morphological changes of the precipitation echoes with substantial fidelity, outperforming the other baselines.

\emph{CNN in Consistency Experiments.---}To minimize the influence of model complexity, a minimal CNN architecture suitable for processing multi-dimensional data was adopted.
This configuration, consisting of two convolutional layers with $16$ and $1$ feature maps respectively, used a kernel size, stride, and padding of $3$, $1$, and $1$.
The simplification of architectural factors enabled a more precise assessment of the performance of each loss function.

\emph{Discussion of AT Loss in Operational Models.---}The AT loss alone is practically sufficient for weather forecasters to distinguish precipitation events greater than or equal to operational thresholds.
However, its use in conjunction with perceptual or adversarial losses is advisable to enhance the details of weather charts~\cite{johnson2016perceptual, goodfellow2014generative}.
Even in the ablation studies, the AT loss was integrated into PCT-CycleGAN along with multiple existing loss functions, including the adversarial loss.
By heuristically optimizing the ratio between AT loss and other losses, this integration produced more realistic forecasts, effective for both light and heavy precipitation events.

Although meteorological agencies worldwide use different models, any gradient-based ML forecasting system can improve its performance through the application of the AT loss. 
Since different ratios among loss functions can yield varying outcomes, it is important to find an appropriate balance with existing losses.
While assigning greater weight to the AT loss offers advantages in forecast verification, relying on it exclusively may reduce sensitivity to detailed values distant from the operational threshold.
Therefore, empirical strategies such as adjusting loss weights, adopting staged training, and applying fine-tuning are essential for achieving optimal forecasts.

\end{document}